\documentclass[10pt,twocolumn,letterpaper]{article}

\usepackage{iccv}              

\definecolor{iccvblue}{rgb}{0.21,0.49,0.74}
\usepackage[pagebackref,breaklinks,colorlinks,allcolors=iccvblue]{hyperref}

\title{3D Scene-Camera Representation with Joint Camera Photometric Optimization}

\author{Weichen Dai$^{1}$ Kangcheng Ma$^{1}$, Jiaxin Wang$^{1}$, Kecen Pan$^{1}$, \\
Yuhang Ming$^{1}$, Hua Zhang$^{1}$, and Wanzeng Kong$^{1*}$
\\
$^{1}$Hangzhou Dianzi University
}

\begin{document}

\twocolumn[{%
\renewcommand\twocolumn[1][]{#1}%
\maketitle
\begin{center}
    \centering
    \captionsetup{type=figure}
    \includegraphics[width=\textwidth,height=5cm]{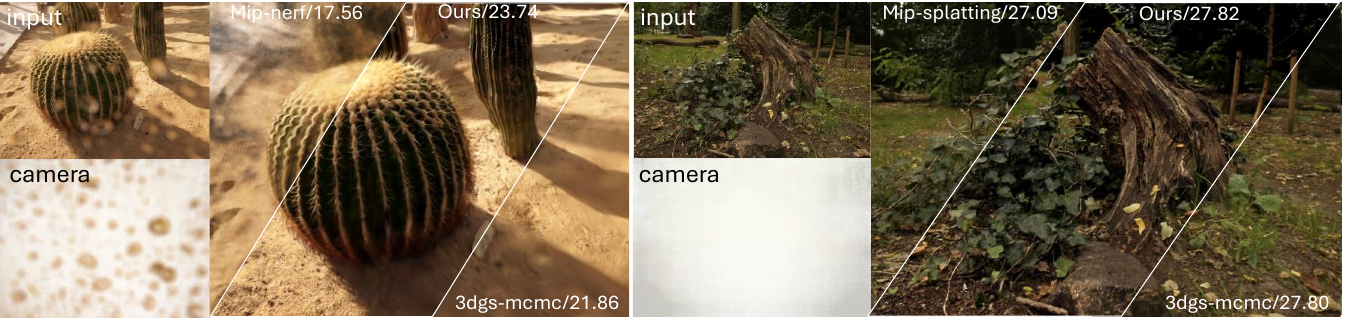}
    \captionof{figure}{\textbf{Comparison of different methods using the images with photometric distortions.} 
    The input images contain severe photometric distortions. Our method effectively separates these distortions (The left image is affected by water droplets, and the right image by vignetting), leading to a more accurate reconstruction with higher fidelity. In each group, the bottom-left image visualizes the photometric distortion modeled by our camera model, the right-side images show results from different methods, with corresponding model names and PSNR values. }
    \label{fig:head}
\end{center}%
}]

\begin{abstract}
Representing scenes from multi-view images is a crucial task in computer vision with extensive applications. However, inherent photometric distortions in the camera imaging can significantly degrade image quality. Without accounting for these distortions, the 3D scene representation may inadvertently incorporate erroneous information unrelated to the scene, diminishing the quality of the representation.
In this paper, we propose a novel 3D scene-camera representation with joint camera photometric optimization. 
By introducing internal and external photometric model, we propose a full photometric model and corresponding camera representation.
Based on simultaneously optimizing the parameters of the camera representation, the proposed method effectively separates scene-unrelated information from the 3D scene representation. Additionally, during the optimization of the photometric parameters, we introduce a depth regularization to prevent the 3D scene representation from fitting scene-unrelated information. By incorporating the camera model as part of the mapping process, the proposed method constructs a complete map that includes both the scene radiance field and the camera photometric model.
Experimental results demonstrate that the proposed method can achieve high-quality 3D scene representations, even under conditions of imaging degradation, such as vignetting and dirt.

\end{abstract}    

\section{Introduction}
\label{sec:intro}
Representing the scene from multi-view 2D images is one of the crucial tasks in computer vision~\cite{schonberger2016structure,mildenhall2020nerf}, with extensive applications in visual effects~\cite{yuan2022nerf}, mixed reality~\cite{li2022rt}, and robotics~\cite{zhou2023nerf,teed2024deep}. Recently, radiance fields—such as NeRF~\cite{mildenhall2020nerf}, 3DGS~\cite{kerbl20233d}, and their extensions~\cite{ren2024nerf,lee2024compact}—have achieved remarkable success in capturing scene and view-dependent effects. 

Numerous subsequent methods~\cite{li2024gp,mildenhall2022nerf,pumarola2021d,lin2021barf,gao2023adaptive} have advanced the performance of radiance fields in various dimensions. These methods are designed to operate on, and are evaluated with, data captured by commodity cameras without taking advantage of knowing – or even being able to influence – the full image formation pipeline, overlooking the variations in light attenuation across different sensor positions due to distortion and contamination introduced by actual camera~\cite{kim2008robust,gu2009removing}. Typically, in addition to camera geometric model~\cite{lin2021barf}, these methods recover the radiance fields without considering the photometric modeling of the camera itself. When photometric distortions are induced by the factors such as lens smudges, lenses with high distortion, and non-uniform response, the imaging quality of the camera is compromised~\cite{li2021let}, which degrades the performance of the radiance fields, as shown in Fig.~\ref{fig:head}.

Fixed photometric distortions related to the camera in images may not significantly impact classic Structure-from-Motion (SfM) methods~\cite{schonberger2016structure}, which primarily rely on modern keypoint detectors and descriptors~\cite{lowe1999object} that are robust or invariant to arbitrary monotonic brightness changes. However, while radiance fields take light variance across different views into account and do not fully assume photometric consistency, the model capacity to accurately fit both scene radiance information and exclude the camera photometric distortions is limited. This limitation create challenges when fitting photometric values with adverse distortions, leading to erroneous geometric estimation and introducing incorrect radiance values.
Therefore, radiance field methods should consider not only the geometric parameters~\cite{lin2021barf} of the camera but also its photometric characteristics.

Utilizing image restoration~\cite{zamir2022restormer,liang2021swinir,li2021let} techniques to address the adverse photometric distortions may present another approach. However, most existing methods focus primarily on single-frame images, relying on prior knowledge to identify defect areas and enhance them. Unfortunately, the supplementary scene information by the image restoration often leads to multi-view images that do not maintain geometric consistency, thereby undermining the foundational principles of radiance field representation.

In this paper, we propose a 3D scene-camera representation method with joint camera photometric optimization that incorporates a comprehensive camera photometric model. The proposed method addresses both internal and external sources of camera photometric distortion, including sensor non-uniform response, lens vignetting, and imperfections such as smudges, thereby considering a full imaging formation. For external sources of camera photometric distortion, we refer to the depth-of-field(DoF) model~\cite{guan2005stress} to reduce the impact of defocus. Based on the photometric model, a camera photometric representation method is designed using a shallow MLP to effectively capture and store camera photometric information.
We optimize the camera photometric parameters and the radiance field of 3D Gaussians alternately, fixing one component while optimizing the other. 
To prevent convergence to local minima and to handle the limited constraints on photometric parameters in sparse-view settings, we further introduce a depth regularization. This regularization applies Gaussian constraints to the opacity distribution during rendering to optimize camera parameters, ensuring that the camera representation does not capture extraneous information.

The main contributions of this work are as follows:
\begin{itemize}
    \item We propose a 3D scene-camera representation method to estimate scene radiance fields and the camera photometric parameters, considering a full imaging formation.
    \item We propose a photometric representation that integrates defocus effects, modeling camera photometric parameters to enable the synthesis of high-quality images free from camera photometric distortions.
    \item By modeling camera photometric representation, the proposed method facilitates the generation of high-fidelity images, demonstrating robustness across both a custom dataset -designed to capture diverse photometric distortions from multiple camera types - and real-world public datasets.
\end{itemize}

\section{Related Work}

\subsection{3D Scene Representation} 
Various representations, including point clouds~\cite{mur2015orb}, meshes~\cite{pan2019deep}, and radiance fields~\cite{fridovich2022plenoxels}, are employed to reconstruct 3D scenes from 2D images for a range of tasks, such as Novel View Synthesis (NVS)~\cite{heigl1999plenoptic, levoy1996light}, SLAM~\cite{cadena2016past}, and Structure-from-Motion (SfM)~\cite{schonberger2016structure}. Recently, neural radiance field (NeRF) methods~\cite{mildenhall2022nerf, barron2022mip, muller2022instant, mildenhall2020nerf} and 3DGS~\cite{kerbl20233d} have demonstrated significant advances in learning scene radiance fields, achieving high-fidelity representations.

For camera geometric parameters, to reduce reliance on prior geometric camera parameters, many existing methods aim to minimize dependence on SfM by integrating pose estimation~\cite{lin2021barf, bian2023nope, fu2024colmapfree3dgaussiansplatting}, similar to SLAM~\cite{cadena2016past}.
However, most of these methods focus exclusively on geometric parameters. Without considering the photometric parameters within the full image formation pipeline, these methods may suffer from reduced performance in scenarios where images are affected by photometric degradation, such as lens attenuation (vignetting) and lens smudges or scratch. This issue is particularly relevant, as data used for radiance field reconstruction is often captured by consumer-grade cameras, which typically have limited control over camera conditions in real-world scenarios.

Existing radiance field representation methods have preliminarily considered photometric factors involved in camera imaging. Some works~\cite{mildenhall2022nerf} utilize raw data captured in low-light conditions to normalize noise, subsequently employing multi-view information to reduce this noise. Some radiance field-based SLAM methods~\cite{matsuki2024gaussian} simplify the radiance fields by adjusting them for greater photometric consistency through isotropic regularization and by employing affine brightness parameters to account for varying exposure~\cite{engel2017direct}. However, few methods currently attempt to represent the camera photometric model alongside the scene from multi-view images, aiming to mitigate the adverse photometric effects of suboptimal camera conditions.

\subsection{Camera Photometric Parameters in 3D Tasks}

The issue of neglecting photometric parameters closely resembles the challenges encountered by direct methods~\cite{engel2017direct, engel2015large} in SLAM. Direct methods define an objective function based on the assumption of photometric consistency~\cite{furukawa2015multi}. By minimizing this photometric loss, which is influenced by each pixel, these methods estimate the poses of the camera and the 3D structure~\cite{schops2019bad, engel2017direct}. However, photometric inconsistencies introduced by the camera can reduce the accuracy and robustness of these methods, leading direct methods to often incorporate photometric parameters.
Similarly, SLAM methods based on the radiance field~\cite{matsuki2024gaussian} often simplify the radiance fields by adjusting them for improved photometric consistency through isotropic regularization and by employing affine brightness parameters to account for varying exposure. Other methods~\cite{mildenhall2022nerf} have recognized the importance of photometric imaging and propose utilizing the raw format to eliminate noise under low light conditions. Nevertheless, current methods for estimating a camera photometric model during radiance field construction remain preliminary.

\subsection{Image Artifact Removal}

Enhancing image quality and removing artifacts is a fundamental task in computer vision~\cite{banham1997digital}, such as image restoration~\cite{liang2021swinir,li2021let,li2022all}, artifact removal~\cite{zamir2022restormer}, and vignetting removal~\cite{luo2024devignet}. These methods typically learn scene texture priors from large datasets and extract features to reconstruct or complete images to the desired quality. However, these techniques are primarily designed for single-view image correction, which can lead to inconsistencies in multi-view tasks and ultimately impact the outcomes of 3D representation models. Additionally, network-based approaches may perform suboptimally when dealing with unseen scenes and objects, resulting in less effective image restoration.

\section{3D Scene-Camera Representation}

The proposed 3D scene-camera representation method is based on the separation of photometric parameters in multi-view images, taking into account a camera photometric model for full image formation pipline. Compared to existing radiance field methods, the proposed method jointly estimates both the scene radiance field and camera photometric parameters. 

\subsection{Camera Photometric Model}
\begin{figure}[t]
    \centering
    \includegraphics[width=\linewidth]{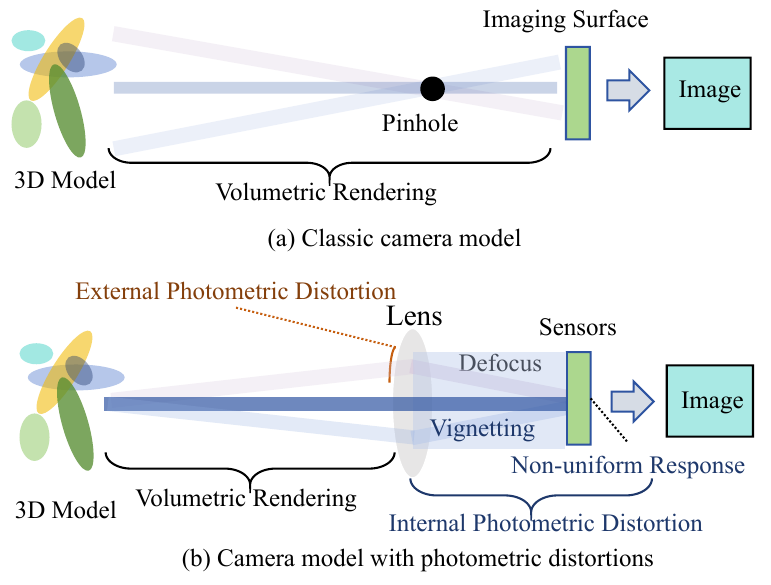}
    \vspace{-20pt}
    \caption{Comparison of image formation pipeline models. }
    \label{fig:camera_model}
\end{figure}

Regarding the camera photometric model, some research has been explored in previous work~\cite{engel2017direct}. As illustrated in Fig.~\ref{fig:camera_model}, existing radiance field rendering methods typically assume an ideal pinhole camera model, where light emitted from a point in space travels in a straight line and directly projects onto the sensor imaging plane to form an image. However, in a complete image formation pipeline, light passes through various components, including the lens and image sensor, each of which affects the final imaging outcome. Factors such as lens smudges, lens attenuation (vignetting), and non-uniform sensor response commonly degrade image quality, introducing photometric distortions.

\begin{figure}[t]
    \centering
    \includegraphics[width=\linewidth]{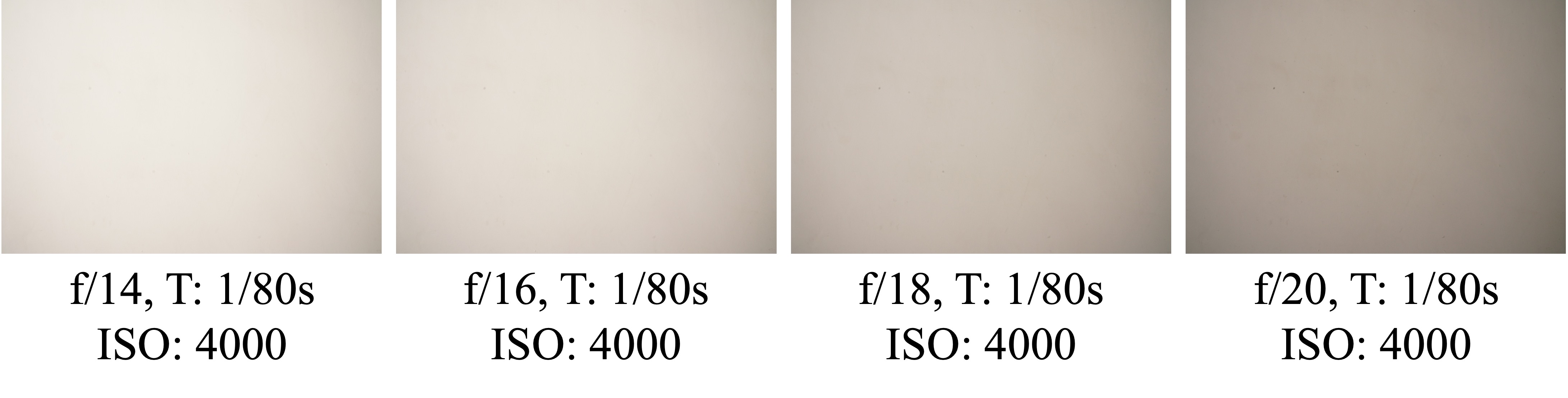}
    \vspace{-15pt}
    \caption{Vignetting under different aperture settings. 
    It is evident that the vignetting in the image becomes more pronounced without good calibration.
    }
    \label{fig:vignettingwithaperture}
\end{figure}

Although most cameras undergo factory calibration to compensate for inherent photometric distortions through the Image Signal Processor~(ISP), changes in camera condition over time can introduce parameter uncertainties, as shown in Fig.~\ref{fig:vignettingwithaperture}. This issue is further exacerbated in real-world scenarios where lens surfaces can accumulate smudges, a common occurrence with cameras.

The proposed method provides a more comprehensive model of the image formation process for 3D scene-camera reconstruction. The geometric camera model consists of a function that projects a 3D point onto a 2D image:
\begin{equation}
    \mathbf{x} = \pi (\mathbf{p})
    \label{equ:projection}
\end{equation}
where \(\mathbf{p}\) denotes the 3D position of the observed point, and \(\pi\) is the projection function that maps \(\mathbf{p}\) to pixel coordinates \(\mathbf{x}\). To simplify the geometric model, we employ the pinhole camera model. With the maturity and widespread use of camera calibration algorithms, radial distortion can be corrected in a preprocessing step, thus allowing us to define \(\pi : \mathbb{R}^3 \rightarrow \mathbb{R}^2\).

Although a pixel on the sensor receives light that has been refracted through the entire lens, the geometry can be simplified to a model where a straight line projects a 3D point onto a 2D image when the scene is in focus. Therefore, inspired by~\cite{engel2016photometrically},  it is also beneficial to consider a photometric camera model, which includes a function that maps the real-world radiance omitted by a position in front of the lens to its corresponding intensity value. Since photometric distortions may arise both within the camera and outside it, the introduction of the camera photometric model is divided into two components: the internal photometric model and the external photometric model.

\subsubsection{Internal Photometric Model}

The main factors contributing to photometric distortions within the camera are vignetting and sensor non-uniformity. Building upon two factors, a photometric camera model that incorporates lens attenuation (vignetting) $V: \mathbb{R}^2 \rightarrow [0, 1]$ and a response $G: \mathbb{R}^2 \rightarrow [0, 1]$.
\begin{equation}
\mathbf{I}_i(\mathbf{x}) = G(\mathbf{x})V(\mathbf{x}) \mathbf{B}_i(\mathbf{x})
\label{equ:internalmodel}
\end{equation}
where $\mathbf{B}_i$ and $\mathbf{I}_i$ are the radiance getting inside the camera and the observed pixel intensity in image $i$. It should be noted that the output of the response function is set to range from 0 to 1 because the scalar of the radiance entering the camera is unknown. Thus, $G$ mapping this value to the interval $[0, 1]$ represents the extent of the radiance response from full transformation to no transformation. Both $V$ and $G$ are functions of pixel coordinates, reflecting the variations in lens and sensor responses across different pixels.

\subsubsection{External Photometric Model}

Camera lenses can accumulate dust, smudges, or other contaminants over time and may also be affected by sudden issues such as lens fogging or obstructions on the lens surface, all of which can introduce distortion and degrade image quality. These surface distortions can manifest as irregular patterns or local variations in image intensity. This can be considered as part of the scene radiance information being partially lost while passing through the contaminants, with the contaminants themselves emitting additional radiance into the camera. Based on this model, we incorporate surface distortion parameters \( S_\alpha \) and \( S_\beta \) into the camera model. The external photometric model can be expressed as:
\begin{equation}
    \mathbf{\hat{B}}_i(\mathbf{x}) = S_\alpha(\mathbf{x}) \mathbf{R}(\mathbf{x}) + S_\beta(\mathbf{x})
    \label{equ:externalmodel}
\end{equation}
where \( S_\alpha: \mathbb{R}^2 \rightarrow [0,1] \) represents the attenuation factor of the contaminants, and \( S_\beta: \mathbb{R}^2 \rightarrow \mathbb{R} \) models the radiance emitted by the contaminants themselves. $\mathbf{R}$ denotes the radiative information emitted by spatial points $\mathbf{p}$ to $\mathbf{x}$. Therefore, the external photometric model captures the impact of photometric distortions on the scene radiance as received by the camera. 

\subsubsection{Defocus Model}
Since real-world images often have finite depth-of-field (DoF), and external sources of camera photometric distortion not on the focal plane, each point not on the focal plane is imaged as circular regions on the sensor plane, instead of single points. The physical model of imaging and DoF has been well studied in geometric optics~\cite{guan2005stress}. Each point $\mathbf{p}$ with the distance from point to lens (object distance) $h_t$ is projected to a circular region (CoC) with centre $\mathbf{x}$ in the imaging plane. The radius of the region ($R_{CoC}$) can be determined by the focal length $f$, the diameter of the aperture $D$ and the focus distance $F$. The calculation formula for the radius of CoC is:
\begin{equation}
R_{CoC} = \frac{1}{2} fD \times \frac{\left|h_t - F\right|}{h_t(F - f)}
\label{equ:radiusOfCoC}
\end{equation}
Since the external sources distortion in our work is mainly on the surface of the lens, these distortions can be regarded as a constant value of $h_t$. And since $D$, $F$, and $f$ are constants during the production of a dataset, it can be regarded that $R_{CoC}$ is a constant value for the lens surface distortion. Within the circular region, the color weight of this point will be uniformly distributed. 
\begin{equation}
\begin{aligned}
    \mathbf{B}_i(\mathbf{x})
    &= \sum_{\mathbf{\hat{x}} \in \text{CoC}_{\mathbf{p}}} \frac{S_\alpha(\mathbf{\hat{x}}) \mathbf{R}(\mathbf{\hat{x}}) + S_\beta(\mathbf{\hat{x}})}{\pi R_{CoC}^{2}} \\
    &= \sum_{\mathbf{\hat{x}} \in \text{CoC}_{\mathbf{p}}} \frac{\mathbf{\hat{B}}_i(\mathbf{\hat{x}})}{\pi R_{CoC}^{2}}
\label{equ:BxAfterSummation}
\end{aligned}
\end{equation}
where $\mathbf{\hat{x}} \in \text{CoC}_{\mathbf{p}}$ indicates that $\mathbf{\hat{x}}$ is within the CoC of point $\mathbf{p}$. 
It can be observed that this result is equivalent to applying a mean convolution operation over a circular region to the camera photometric distortion. 

\subsubsection{Full Photometric Model}
Combining internal and external photometric model Equ.~\eqref{equ:internalmodel}, \eqref{equ:externalmodel}, and \eqref{equ:BxAfterSummation}, the full photometric model under defocus can be conducted. 
\begin{equation}
\begin{aligned}
    \mathbf{I}_i(\mathbf{x}) 
    &= G(\mathbf{x})V(\mathbf{x}) \sum_{\mathbf{\hat{x}} \in \text{CoC}_{\mathbf{p}}} \frac{S_\alpha(\mathbf{\hat{x}}) \mathbf{R}(\mathbf{\hat{x}}) + S_\beta(\mathbf{\hat{x}})}{\pi R_{CoC}^{2}} \\
    & = M(\mathbf{x}) \sum_{\mathbf{\hat{x}} \in \text{CoC}_{\mathbf{p}}} \frac{S_\alpha(\mathbf{\hat{x}}) \mathbf{R}(\mathbf{\hat{x}}) + S_\beta(\mathbf{\hat{x}})}{\pi R_{CoC}^{2}} 
\label{equ:fullmodel}
\end{aligned}
\end{equation}
where \( M(\mathbf{x}) = G(\mathbf{x}) V(\mathbf{x}) \) represents the integrated attenuation factors. Since \( G(\cdot) \), \( V(\cdot) \)and \( S_\alpha(\cdot) \)  all map to values between 0 and 1, \( M(\cdot): \mathbb{R}^2 \rightarrow [0,1]  \) remains in the range of 0 to 1.

\begin{equation}
\begin{aligned}
    \mathbf{I}_i(\mathbf{x})
    &=M(\mathbf{x}) (S_\alpha(\mathbf{{x}}) \mathbf{R}(\mathbf{{x}}) + S_\beta(\mathbf{{x}}))
\label{equ:simplemodel}
\end{aligned}
\end{equation}
It should be noted that the points in the scene are considered to be within the DoF, so Equ.~\eqref{equ:simplemodel} can be used to simplify the processing.

\begin{figure*}
    \centering
    \includegraphics[width=\linewidth]{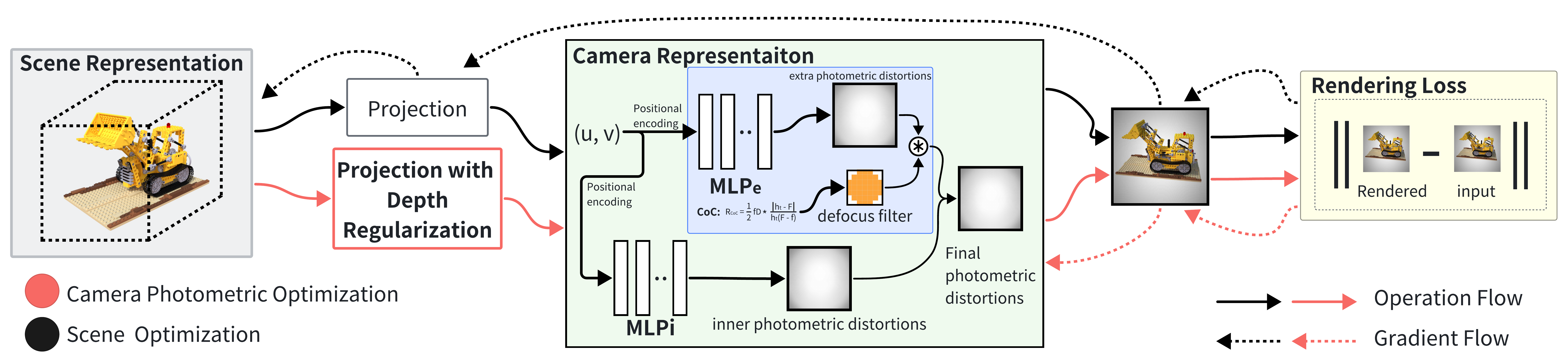}
    \caption{System overview. In the figure, red arrows indicate camera representation training and black arrows indicate scene representation training. We first optimize the camera photometric representation using projection with depth regularization to reduce the influence of floating artifacts on the camera representation, then optimize the scene representation while keeping the camera model fixed.}
    \label{fig:overview}
\end{figure*}

\subsection{Representation Method}

The overview of the proposed method is illustrated in Fig.~\ref{fig:overview}. The core of the proposed method lies in incorporating a camera representation model into the dense 3D Gaussian representation. We need to mention that $u$ and $v$ in the figure are the two-dimensional coordinate representations of $x$ in Equ.~\eqref{equ:projection}.

\subsubsection{Scene Representation}

The scene is represented as a set of 3D Gaussians~\cite{kerbl20233d}, an explicit form of a radiance field representation. Each 3D Gaussian is characterized by a covariance matrix \( \boldsymbol{\Sigma} \) and a center point \( \boldsymbol{\mu} \) in the 3D space.
As introduced in~\cite{kerbl20233d}, the rendering process from a given camera view involves splatting the Gaussian onto the image plane. This is achieved by approximating the projection of a 3D Gaussian along the depth dimension into pixel coordinates. 
The final rendered color can be formulated as the alpha-blending of N ordered points that overlap the pixel, 
\begin{equation}
    \mathbf{R}(\mathbf{x}) = \sum_n^N  c_n \alpha_n \prod^{n-1}_m (1-\alpha_m)
    \label{equ:scenemodel}
\end{equation}
where $c_n$, $\alpha_n$ represents the color and density of this point computed from the learnable per-point opacity and SH color coefficients weighted of $n$-th Gaussian.
Through this splatting rendering technique, all parameters of the 3D Gaussian, along with spherical harmonics (SH) and opacity, can be efficiently optimized using the gradient from the loss. It should be noted that our work uses 3DGS-mcmc\cite{kheradmand20243dmcmc} as the baseline for Scene Representation.

\subsubsection{Camera Representation}
\label{sec:camerarep}
In practical camera imaging, the camera primarily focuses on the scene rather than on stains on the lens surface, meaning that the primary attention is on the subject being captured. As a result, stains on the lens typically appear blurred. Additionally, vignetting in the lens and the non-uniform response of the image sensor are generally smooth in nature. Therefore, the photometric parameters of adjacent pixels are correlated. In the proposed method, based on Equ.~\eqref{equ:fullmodel}, the camera photometric parameters will be fitted using a shallow MLP model, which is a compact model correlating all values across the pixel coordinates, 
\begin{equation}
    \mathbf{I}_i(\mathbf{x})
    = mlp_\alpha(\mathbf{x}) \sum_{\mathbf{\hat{x}} \in \text{CoC}_{\mathbf{p}}} \frac{mlp_\beta(\mathbf{\hat{x}}) \mathbf{R}(\mathbf{\hat{x}}) + mlp_\gamma(\mathbf{\hat{x}})}{\pi R_{CoC}^{2}} 
    \label{equ:cameramodel}
\end{equation}
where $mlp_\alpha$, $mlp_\beta$ and $mlp_\gamma$ are the outputs of the MLP network. It should be noted that $mlp_\alpha$ is the output of the $MLP_i$ representing the internal photometric distortion, while $mlp_\beta$ and $mlp_\gamma$ are the outputs of the $MLP_e$ representing the internal photometric distortion.
In this equation, the various photometric distortion factors affecting the input radiance are modeled through the MLP network for scaling and bias correction, reducing the number of parameters.
Since $R_{CoC}$ is a constant for a given scene, we set it via a hyperparameter.

\subsection{Parameter Optimization}

\subsubsection{Loss Function}

During the optimization process, both the camera photometric parameters and 3D Gaussian Scene (3DGS) parameters are optimized using two types of loss functions. 

First, by combining Equ.~\eqref{equ:scenemodel} and \eqref{equ:cameramodel}, the final rendered image can be obtained. This enables the computation of a loss function based on direct image supervision:
\begin{equation}
    \mathcal{L}_{pho} = (1-\lambda) \| \mathbf{I}_i - \mathbf{I}_i^{input} \|_1 + \lambda \mathcal{L}_{ssim}
\end{equation}
where \( \mathbf{I}^{input} \) denotes the images captured by the camera, and \( \mathcal{L}_{ssim} \) represents the D-SSIM term. $\lambda$ is a hyperparameter that balances two losses.

\subsubsection{Projection with Depth Regularization}
\begin{figure}
    \centering
    \includegraphics[width=\linewidth]{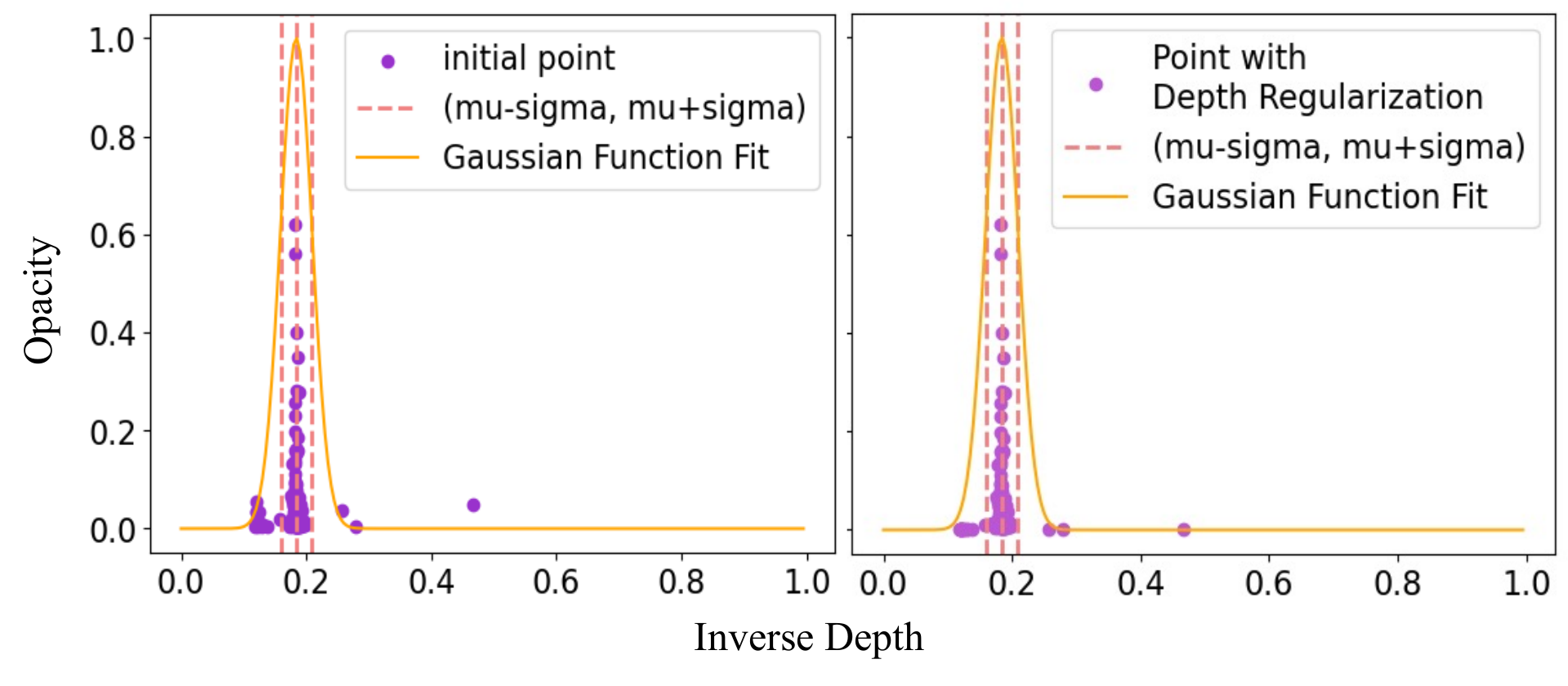}
    \vspace{-10pt}
    \caption{Opacity distribution along one ray. The set of purple points represents the fitted Gaussian distribution, while the pink dashed line indicates the range of one standard deviation on either side. }
    \label{fig:distribution}
\end{figure}

Although the MLP is used to model the photometric parameters, when combined with the scene radiance field representation, the camera photometric representation has an excessive fitting capacity, which sometimes leads to overfitting, particularly with the limited, narrow - perspective data. To mitigate this overfitting issue, we propose a depth regularization technique based on prior knowledge: during the projection process, only the object surface exhibits maximum opacity, as shown in Fig.~\ref{fig:distribution}. Therefore, we reduce the interference from floating points by minimizing the contribution of these overfitting points in the projected images. Our designed Depth Regularization aims to prevent overfitting of the camera photometric representation. It solely impacts the camera representation, not the scene representation.

Since the radiance field is represented by multiple 3D Gaussians (3DGS), the opacity near the object surface during projection should approximately follow a Gaussian distribution. To address this, for each pixel, we compute the inverse depth~\cite{civera2008inverse} of the associated 3DGS based on the mean and variance and then apply a Gaussian function to suppress points lying outside the 1-sigma range. By adjusting the opacity based on the Gaussian function values for optimizing camera photometric parameters, we effectively suppress points that significantly deviate from the object surface, thus mitigating the overfitting issue.
\begin{equation}
    \mathbf{R}(\mathbf{x}) = \sum_n^N  c_n \alpha_n G_{inv}(id_i) \prod^{n-1}_m (1-\alpha_m)
\end{equation}
where 
\begin{equation}
    G_{inv}(id) = e^{-\frac{({id}-{\mu_{id}})^2}{2\sigma_{id}^2}}
\end{equation}
\begin{equation}
    \boldsymbol{\mu}_{id} = \sum_n^N \alpha_n id_{n}
\end{equation}
where $\mu_{id}$ and $\sigma_{id}$ is the mean and standard deviation of the inverse depth of the 3DGS along the line of sight.

\subsubsection{Alternating Camera and Scene Optimization}

The camera model requires multiple views as input for fitting, while the 3D reconstruction model typically uses a single image with specific camera coordinates and angles. Therefore, the optimization process adopts an iterative training approach to optimize the camera photometric parameters and the scene photometric parameters sequentially.
For the camera model, we select a set of viewpoints based on the current state of the 3D reconstruction model, project the scene to obtain rendered images, and compute the loss between the final rendered images obtained through the camera model and the corresponding input images. In each iteration, we first update and train the camera model, and then pause the updates to proceed with the training of the 3D reconstruction model.

\section{Experiments}

\subsection{Dataset}

\begin{figure}[t!]
    \centering
    \includegraphics[width=\linewidth]{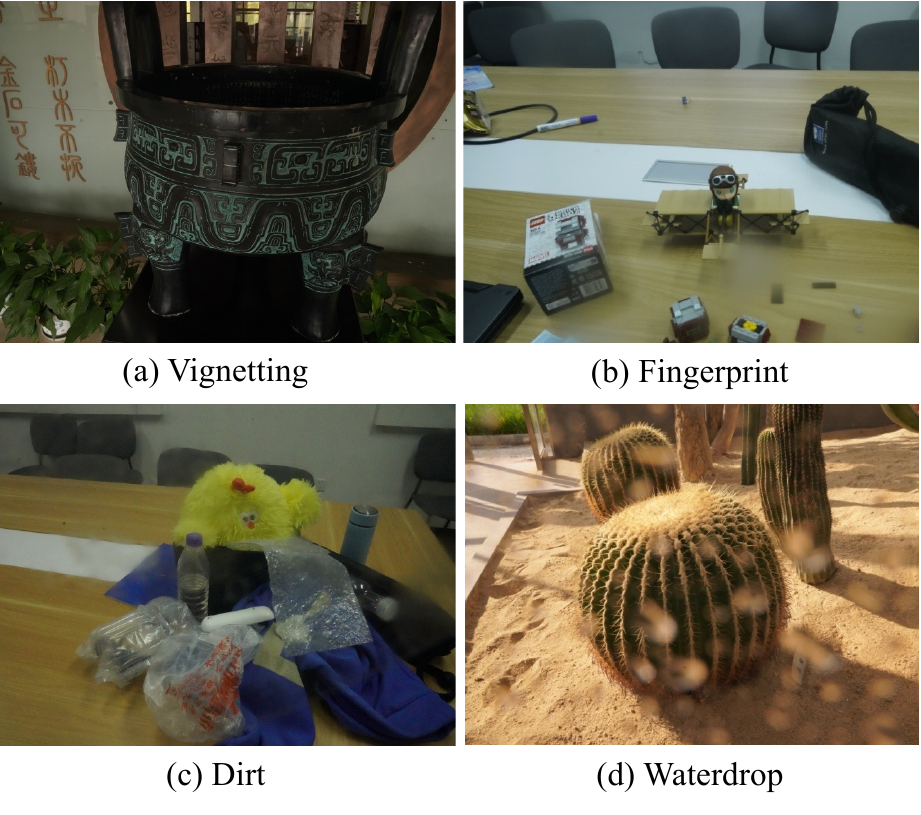}
    \vspace{-20pt}
    \caption{Different mixed photometric distortions}
    \label{fig:realdataset}
\end{figure}

\textbf{Public Dataset:}
We evaluate our camera model's capability in modeling photometric distortions using the MipNeRF360\cite{barron2022mip} dataset, which consists of real-world captures. Additionally, we assess our model on the NeRF-Synthetic\cite{mildenhall2020nerf} dataset to demonstrate that it does not introduce artifacts or degrade performance in scenarios where the camera is free from photometric distortions. \textbf{Our Dataset:} As there are currently no suitable datasets collected under adverse conditions with varying photometric distortions, we additionally used a Sony camera for handheld shooting, capturing between 40 to 60 images of various scenes. This dataset includes different types of photometric distortions, such as camera vignetting, fingerprints on the lens surface, dirt, and water droplets, as shown in Fig.~\ref{fig:realdataset}. These photometric distortions are introduced to assess their impact on rendering quality and evaluate the performance. Please see supplementary materials for more information.

\subsection{Baselines}

We selected a diverse set of baselines for comparative experiments to validate the effectiveness and superiority of our camera model. The baselines are categorized into implicit and explicit representations. Among the implicit representations, we compared against Mip-NeRF~\cite{barron2021mip}, INGP~\cite{muller2022instant}, 
and NeuRBF~\cite{chen2023neurbf}, all of which have been demonstrated to possess strong scene-fitting capabilities. 

In the category of explicit representations, we included not only 3DGS~\cite{kerbl3dgs} but also the latest state-of-the-art scene reconstruction model based on explicit representation, 3DGS-MCMC~\cite{kheradmand20243dmcmc} and Mip-Splatting~\cite{yu2024mip}. The strong baseline can further demonstrate that our model can better represent scenes captured by real cameras, rather than merely enhancing the model's fitting capability.

\subsection{Quantitative Evaluation}

\begin{figure} 
    \centering
    \includegraphics[width=\linewidth]{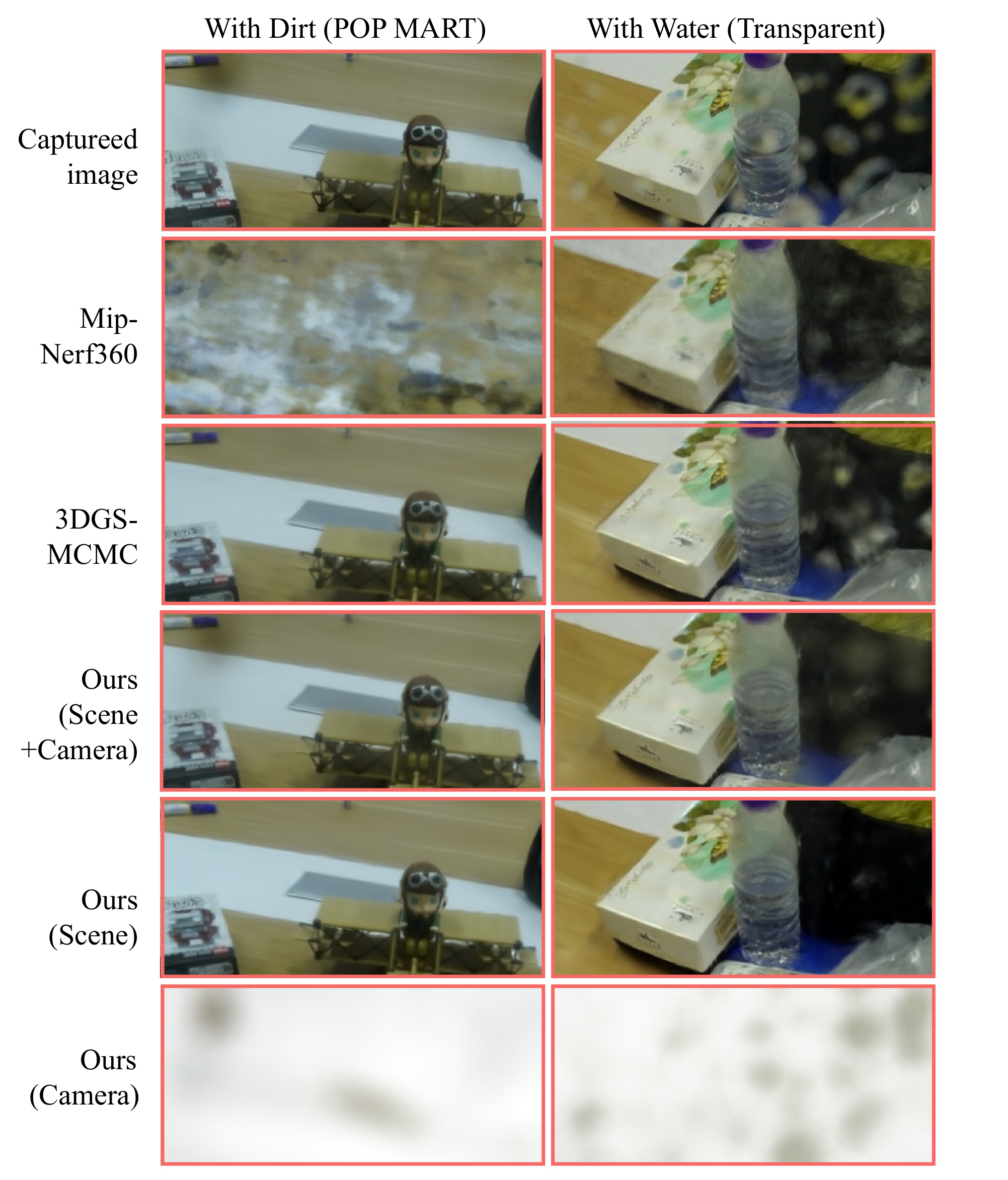}
    \caption{ Comparison of different methods using the images with photometric distortions (dirt and water). }
    \label{fig:mixd_pho_dist}
\end{figure}

We first evaluate our model with our dataset. As shown in Table \ref{tab:diffetent_artifact}, the proposed method outperforms other state-of-the-art methods in the presence of mixed photometric distortions. The internal photometric distortions, such as vignetting, often go unnoticed by human observers. The proposed method effectively mitigates the impact of subtle vignetting on 3D representation, even in scenes where it is imperceptible to the human eye, as shown in Fig.~\ref{fig:head}. This demonstrates the effectiveness of the proposed method and underscores the importance of jointly constructing the 3D radiance field with the camera photometric model in practice. Notably, the proposed method shows a greater improvement in scenes with the remaining three types of mixed photometric distortions compared to vignetting, as shown in Fig.~\ref{fig:mixd_pho_dist}. This proves that as the impact of photometric distortions increases, the methods that do not account for the camera photometric model become less effective.
Additionally, the performance of implicit representations indicates that such methods, compared to explicit models like 3DGS, are more sensitive to these distortions and tend to fit noise in the scene more easily.

\begin{table}[!t]
    \centering
    \caption{Comparison results across different photometric distortions.}
    \resizebox{\linewidth}{!}
    {
    \begin{tabular}{@{}lcc|cc|cc|cc@{}}
    \toprule
    \textbf{}  & \multicolumn{2}{c}{\textbf{Fingerprint}} & \multicolumn{2}{c}{\textbf{Dirt}} & \multicolumn{2}{c}{\textbf{water droplet}} & \multicolumn{2}{c}{\textbf{vignetting}}\\
    \midrule
    \textbf{Method}  & \textbf{PSNR} & \textbf{SSIM} & \textbf{PSNR} & \textbf{SSIM} & \textbf{PSNR} & \textbf{SSIM} & \textbf{PSNR} & \textbf{SSIM}\\
    \midrule
    Mip-NeRF & 21.36 & 0.67 & 20.56 & 0.65 & 19.88 & 0.58 & 20.74 & 0.60\\
    INGP  & 22.34 & 0.70 & 21.87 & 0.67 & 20.15 & 0.62 & 21.28 & 0.63\\
    NeuRBF & 20.67 & 0.64 & 21.95 & 0.67 & 19.99 & 0.59 & 20.50 & 0.58\\
    Mip-Splatting & 25.82 & 0.80 & 24.45 & 0.77 & 22.21 & 0.697 & 24.20 & 0.74\\
    3DGS & 26.43 & 0.82 & 25.38 & 0.80 & 23.68 & 0.74 & 24.72 & 0.75\\
    3DGS-MCMC & \colorbox{orange!50}{27.81} & \colorbox{orange!50}{0.84} & \colorbox{orange!50}{26.50} & \colorbox{orange!50}{0.82} & \colorbox{orange!50}{24.64} & \colorbox{orange!50}{0.76} & \colorbox{orange!50}{24.72} & \colorbox{orange!50}{0.76}\\
    Ours  & \colorbox{red!50}{28.91} & \colorbox{red!50}{0.85} & \colorbox{red!50}{27.75} & \colorbox{red!50}{0.83} & \colorbox{red!50}{25.82} & \colorbox{red!50}{0.78} & \colorbox{red!50}{25.35} & \colorbox{red!50}{0.76}\\
    \bottomrule
    \end{tabular}
    }
    \label{tab:diffetent_artifact}
\end{table}

\begin{figure}[!t]
    \centering
    \includegraphics[width=\linewidth]{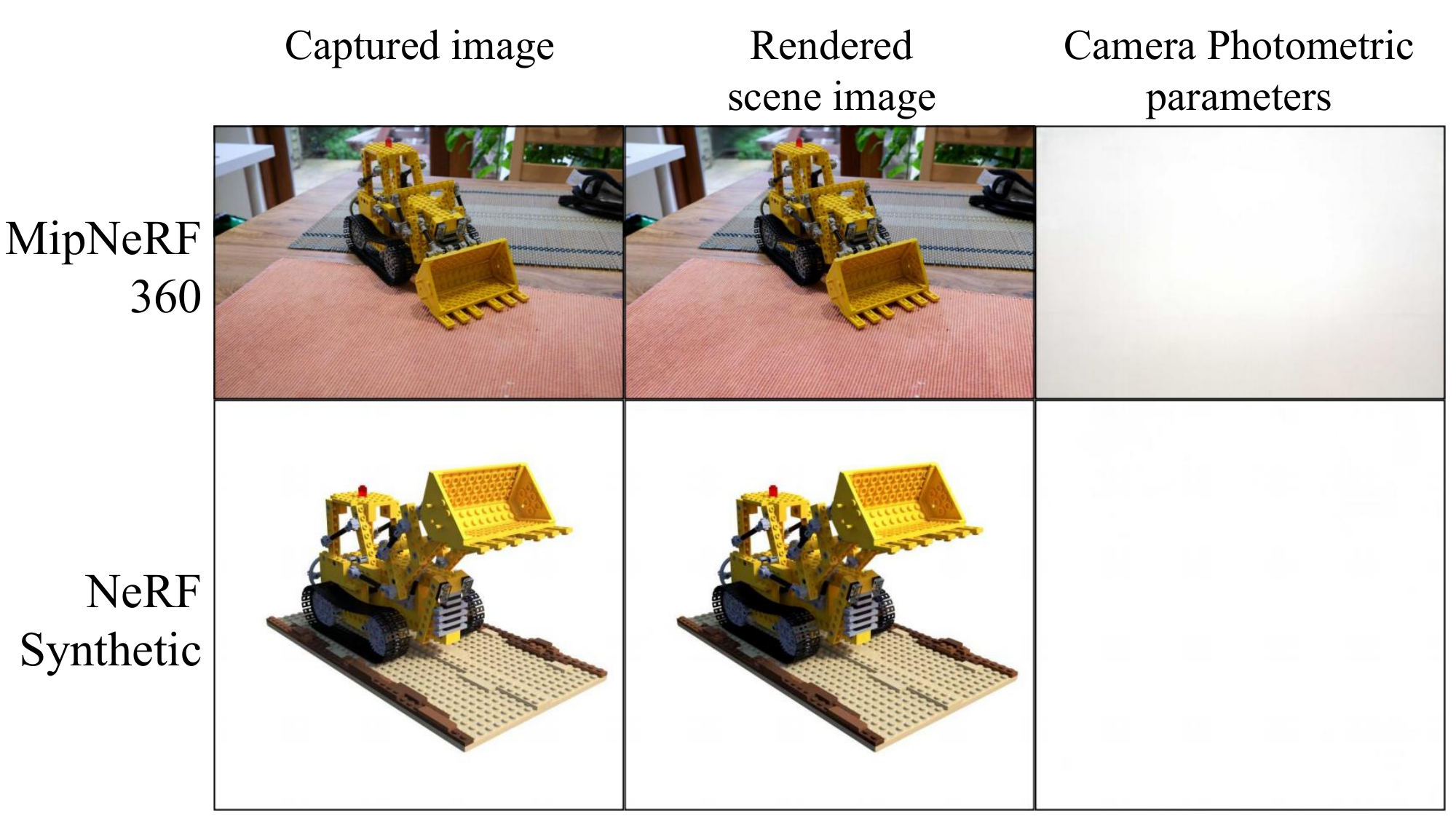}
    \caption{Visualizing the comparison on the public dataset}
    \label{fig:public_comp}
\end{figure}

To further evaluate the performance of the proposed method, we have added experiments using public dataset. Our dataset processing approach aligns with that of our 3DGS-MCMC baseline work.
As shown in the table \ref{tab:public_comp}, our method performs reliably on the public dataset. While Mip-Splatting performs best PSNR on synthetic datasets, our method outperforms it on real data, demonstrating its effectiveness in addressing photometric distortions in real-world scenarios. We also provide visual results in Fig.~\ref{fig:public_comp} to demonstrate the effectiveness of our method, and this also indicates the general nature of the issue we have addressed. 

The data above were all collected from cameras with the same internal photometric distortions. To verify that the proposed method is applicable to different cameras, we collected data from three different cameras, including a Logi Web, a Sony A7R3, and a OPPO reno7, capturing two distinct scenes. As shown in Table~\ref{tab:diffetent_camera}, the proposed method demonstrates a significant performance improvement compared to the 3DGS scene representation model (baseline), demonstrating the robustness of the proposed method to varying internal photometric distortions across different cameras.

\begin{table}[t!]
    \centering
    \caption{ Comparison on the public dataset. }
    \resizebox{\linewidth}{!}
    {
    \begin{tabular}{@{}lccc|ccc|ccc@{}}
    \toprule
    \textbf{} & \multicolumn{3}{c}{$\textbf{Our dataset}$ (Vignetting)} & \multicolumn{3}{c}{\textbf{NeRF Synthetic}} & \multicolumn{3}{c}{\textbf{MipNeRF 360} (Real)}\\
    \midrule
    \textbf{Method} & \textbf{PSNR} & \textbf{SSIM} & \textbf{LPIPS}  & \textbf{PSNR} & \textbf{SSIM} & \textbf{LPIPS} & \textbf{PSNR} & \textbf{SSIM} & \textbf{LPIPS}\\
    \midrule  
    3DGS & 24.72 & 0.75 & 0.27 & 33.42 & 0.97 & 0.04 & 28.69 & 0.87 & 0.22\\
    Mip-Splatting & 24.20 & 0.74 & 0.28 & \colorbox{red!50}{34.25} & \colorbox{orange!50}{0.98} & \colorbox{orange!50}{0.02} & 29.46 & 0.88 & \colorbox{orange!50}{0.16}\\
    3DGS-MCMC & \colorbox{orange!50}{25.18} & \colorbox{orange!50}{0.76} & \colorbox{orange!50}{0.25} & 33.80 & 0.97 & 0.04 & \colorbox{orange!50}{29.89} & \colorbox{orange!50}{0.90} & 0.19\\
    Ours & \colorbox{red!50}{25.35} & \colorbox{red!50}{0.76} & \colorbox{red!50}{0.24} & \colorbox{orange!50}{34.03} & \colorbox{red!50}{0.98} & \colorbox{red!50}{0.02} & \colorbox{red!50}{29.97} & \colorbox{red!50}{0.90} & \colorbox{red!50}{0.15}\\
    \bottomrule
    \end{tabular}
    }
    \label{tab:public_comp}
\end{table}

\begin{table}[!t]
    \centering
    \caption{Comparison across different cameras. Baseline means the 3DGS scene representation model.
    }
    \resizebox{\linewidth}{!}
    {
    \begin{tabular}{@{}lcc|cc|cc@{}}
    \toprule
    \textbf{} & \multicolumn{2}{c}{\textbf{Baseline}} & \multicolumn{2}{c}{\textbf{Ours}} & \multicolumn{2}{c}{\textbf{Improvement}}  \\
    \midrule
    \textbf{Cameras} & \textbf{PSNR} & \textbf{SSIM} & \textbf{PSNR} & \textbf{SSIM} & \textbf{PSNR} & \textbf{SSIM} \\
    \midrule
    Sony & 25.32 & 0.88 & 27.60 & 0.89 & 9.0\%\textcolor{red}{\textuparrow} & 1.6\%\textcolor{red}{\textuparrow} \\
    Logi web camera & 23.98 & 0.91 & 26.42 & 0.93 & 10.2\%\textcolor{red}{\textuparrow} & 2.0\%\textcolor{red}{\textuparrow} \\
    Mobile phone & 22.14 & 0.84 & 24.03 & 0.87 & 8.5\%\textcolor{red}{\textuparrow} & 4.4\%\textcolor{red}{\textuparrow} \\
    \bottomrule
    \end{tabular}
    }
    \vspace{-10pt}
    \label{tab:diffetent_camera}
\end{table}

\subsection{Qualitative Evaluation}

\begin{figure}[t]
    \centering
    \includegraphics[width=\linewidth]{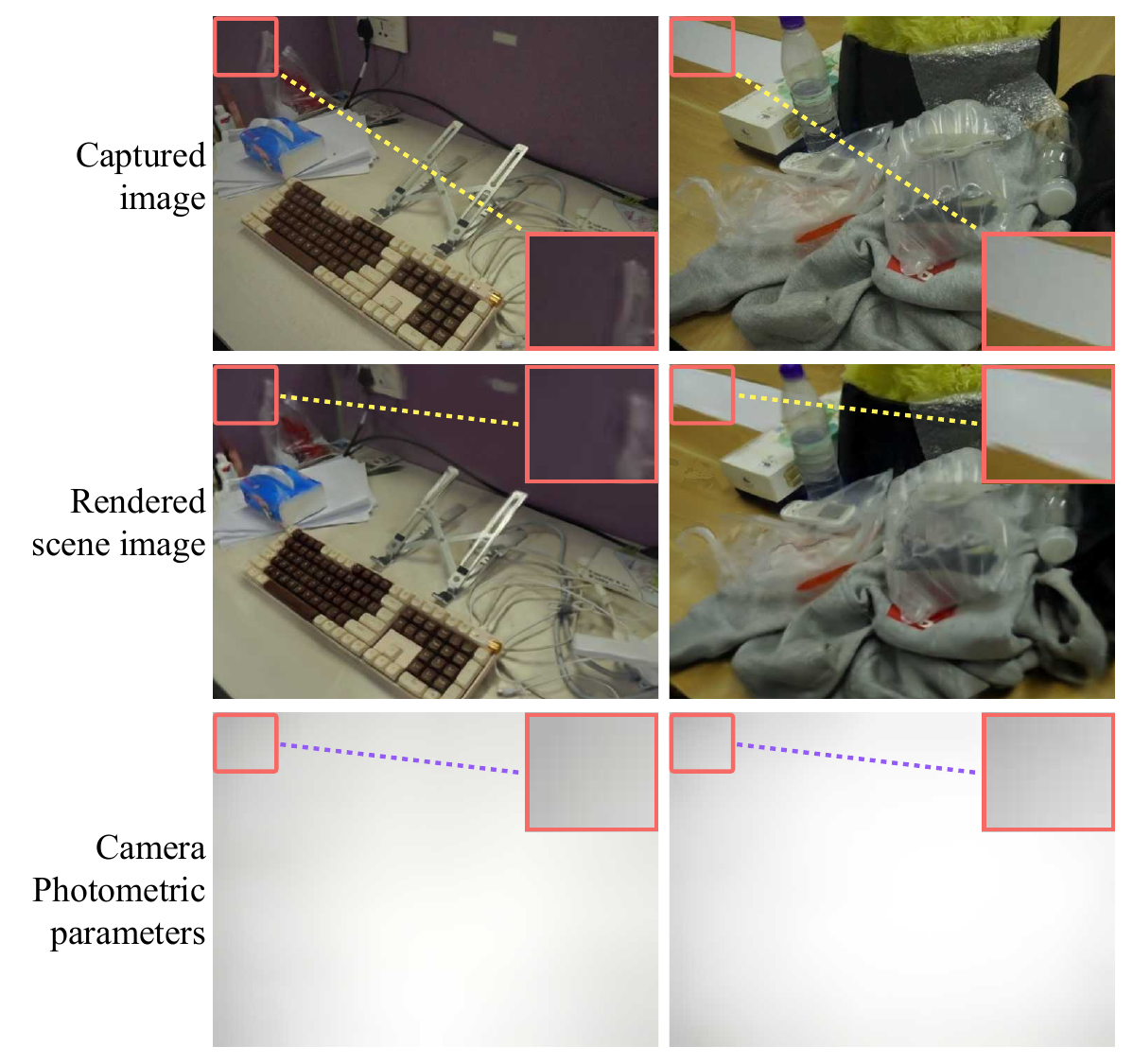}
    \caption{Comparison on different scene images captured with the same camera}
    \label{fig:camera_accuracy}
\end{figure}

To evaluate the effectiveness of the camera photometric parameters extracted by the proposed method, we conducted experiments using data captured with the same camera across two different scenes. The results are shown in Fig.~\ref{fig:camera_accuracy}. By comparing images captured with the actual camera to images only rendered from the scene radiance field, we can observe that the proposed method effectively removes the effects of vignetting. The camera photometric parameters are obtained by inputting an image with uniform maximum pixel values.
The photometric parameter maps of the two different scenes show similar estimated internal photometric parameters of the camera. This qualitative result demonstrates the effectiveness of the proposed method.

\subsection{Ablation Study}

Ablation experiments show that the proposed camera photometric model, defocus model and depth regularization all contribute to the improved performance of the proposed method, as shown in Table~\ref{tab:ablation}. The optimization of the camera photometric representation significantly enhances the overall performance of the proposed method. Explicit modeling of the depth regularization and the defocus model benefits novel view synthesis, and depth regularization and the defocus model can promote each other when they are jointly optimized.

\begin{table}[!t]
    \centering
    \caption{Ablation experiments. CPR is camera photometric representation. DR denotes depth regularization. DM is defocus model.}
    \resizebox{0.8\linewidth}{!}
    {
    \begin{tabular}{@{}l|c|c|c|c|c@{}}
    \toprule
    \textbf{Model} & \textbf{CPR} & \textbf{DR} & \textbf{DM} & \textbf{PSNR(↑)} & \textbf{SSIM(↑)}\\
    \midrule
    baseline & $\times$ & $\times$ & $\times$ & 26.034 & 0.794\\
    ours & $\checkmark$ & $\times$ & $\times$ & 26.631 & 0.801\\
    ours & $\checkmark$ & $\checkmark$ & $\times$ & 26.752 & 0.801\\
    ours & $\checkmark$ & $\times$ & $\checkmark$ & 26.903 & 0.802\\
    ours & $\checkmark$ & $\checkmark$ & $\checkmark$ & 26.955 & 0.803\\
    \bottomrule
    \end{tabular}
    }
    \vspace{-10pt}
    \label{tab:ablation}
\end{table}

\section{Conclusion}

In this paper, we proposed a novel method for 3D scene-camera reconstruction with joint camera photometric optimization. By analyzing the factors contributing to both internal and external photometric distortions, we constructed a model for camera photometric distortion. Based on this model, we designed a photometric representation. By integrating joint parameter optimization of the camera photometric representation, the method effectively addresses photometric distortions, preventing them from negatively impacting the radiance field construction.
The experiments demonstrated that the proposed method outperforms existing state-of-the-art methods, particularly in the presence of mixed photometric distortions such as vignetting combined with fingerprints, dirt, and water droplets. Ablation studies confirmed that all components, including the camera photometric model and depth regularization, contributed to the method performance.
Overall, the proposed method provides a promising approach for more accurate full-map reconstruction, including both scene radiance fields and camera parameters, making it suitable for real-world applications with diverse camera setups and varying conditions.

{
    \small
    \bibliographystyle{ieeenat_fullname}
    \bibliography{main}
}

\end{document}